\pdfoutput=1

\documentclass[11pt]{article}

\usepackage[]{acl}

\usepackage{times}
\usepackage{latexsym}

\usepackage[T1]{fontenc}

\usepackage[utf8]{inputenc}

\usepackage{microtype}

\usepackage{colortbl}
\usepackage{graphicx}
\usepackage{bm}
\usepackage{amsmath}

\usepackage{multirow}
\usepackage{algorithm}
\usepackage{algorithmicx}
\usepackage{algpseudocode} 

\usepackage{algpseudocode}
\usepackage{bbding}
\usepackage{booktabs}

\usepackage{fbox} 
\usepackage{tcolorbox}
\newtcbox{\mybox}[1][red]
  {on line, arc = 0pt, outer arc = 0pt,
    colback = #1!10!white, colframe = #1!50!black,
    boxsep = 0pt, left = 1pt, right = 1pt, top = 1pt, bottom = 1pt,
    boxrule = 0pt, bottomrule = 0.5pt, toprule = 0.5pt}
    
\title{Generating Hierarchical Explanations on Text Classification \\ Without Connecting Rules}

\author{
    Yiming Ju\textsuperscript{},
    Yuanzhe Zhang\textsuperscript{},   
    \textbf{Kang Liu\textsuperscript{},} 
    \textbf{Jun Zhao\textsuperscript{},}
    \\
    \textsuperscript{\rm 1} National Laboratory of Pattern Recognition, Institute of Automation, CAS, Beijing, China \\
    \textsuperscript{\rm 2} School of Artificial Intelligence, University of Chinese Academy of Sciences, Beijing, China \\
    \texttt{\{yiming.ju, yzzhang, kliu, jzhao\}@nlpr.ia.ac.cn} \\
}

\begin{document}
\maketitle
\begin{abstract}
The opaqueness of deep NLP models has motivated the development of methods for interpreting how deep models predict.
Recently, work has introduced hierarchical attribution,
which produces a hierarchical clustering of words, along with an attribution score for each cluster.
However, existing work on hierarchical attribution all follows the connecting rule, limiting the cluster to a continuous span in the input text. We argue that the connecting rule as an additional prior may undermine the ability to reflect the model decision process faithfully.
To this end, we propose to generate hierarchical explanations without the connecting rule and introduce a framework for generating hierarchical clusters.
Experimental results and further analysis show the effectiveness of the proposed method in providing high-quality explanations for reflecting model predicting process.

\end{abstract}

\section{Introduction}
The opaqueness of deep natural language processing (NLP) models has grown in tandem with their power \citep{doshi2017towards}, which has motivated efforts to interpret how these black-box models work \citep{sundararajan2017axiomatic, belinkov2019analysis}.
Post-hoc explanation aims to explain a trained model and reveal how the model arrives at a decision \citep{jacovi2020towards, molnar2020interpretable}. 
In NLP, this goal is usually approached with attribution method, which assesses the influence of inputs on model predictions.

\begin{figure}[htb]
	\centering
	\includegraphics[width=0.95\linewidth]{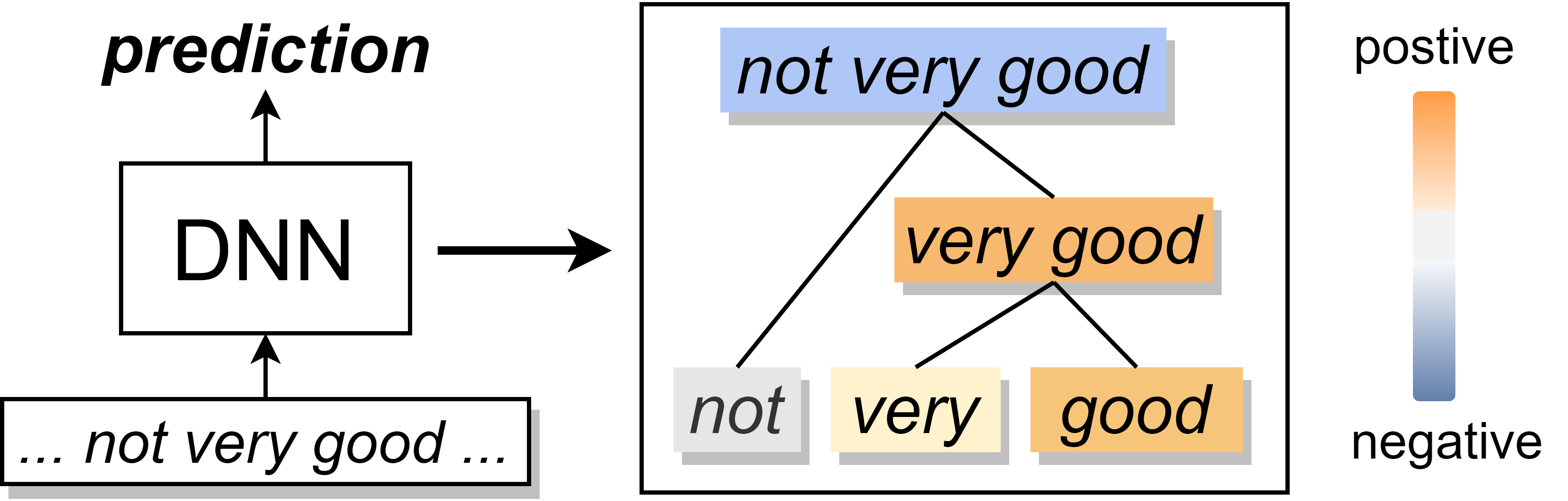}
	\caption{An example of hierarchical attribution.
}
	\label{hierarchical attribution}	
\end{figure}

Prior lines of work on post-hoc explanation usually focus on generating word-level or phrase-level attribution for deep NLP models.
Recently, work has introduced the new idea of hierarchical attribution \citep{singh2018hierarchical, jin2019towards, chen2020generating}.
As shown in Figure \ref{hierarchical attribution},
hierarchical attribution produces a hierarchical clustering of words,  and provides attribution scores for each clusters. By providing compositional semantics information, hierarchical attribution can give users a better understanding of the model decision-making process.
Since the attribution score of each cluster in hierarchical attribution is calculated separately, the key point of generating hierarchical attribution is how to get word clusters,
which should be informative enough to capture meaningful feature interactions while displaying a sufficiently small subset of all feature groups to maintain simplicity \citep{singh2018hierarchical}.

\begin{figure*}[t!]
	\centering
	\includegraphics[width=0.99\linewidth]{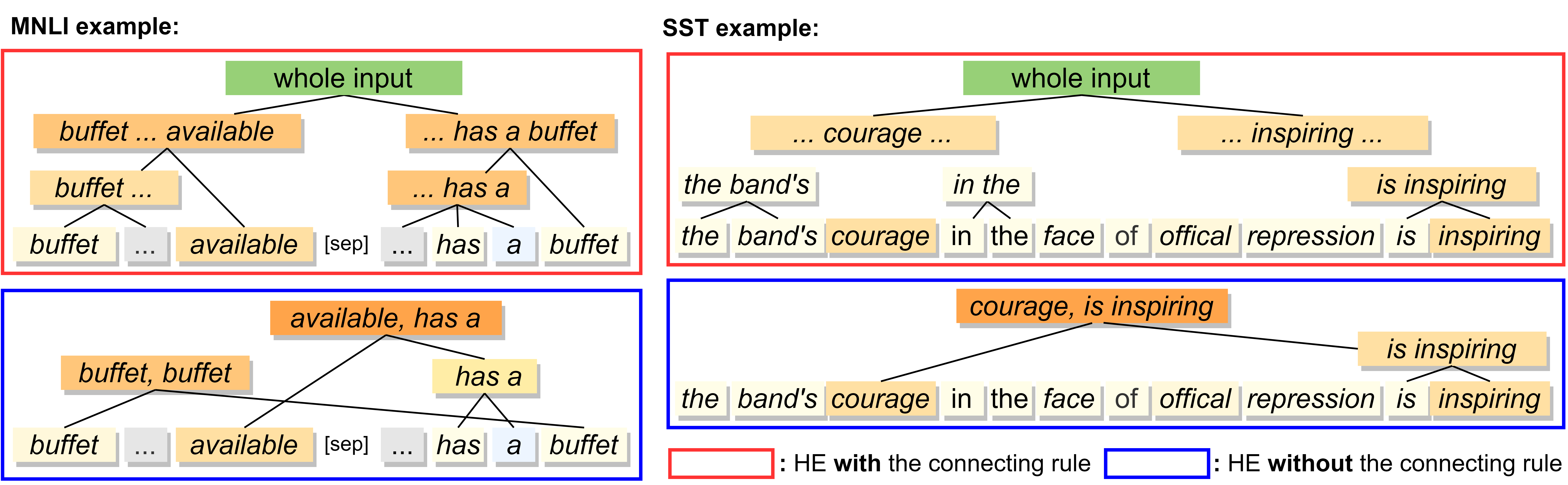}
	\caption{Examples of hierarchical explanations. The prediction of MNLI example is entailment (the first sentence entails the second sentence.)  The prediction of SST example is positive. `...' represent omitted words for clear visualization. }
	\label{our method}	
\end{figure*}

Existing work has proposed various algorithms to generate hierarchical clusters. For example, \citet{singh2018hierarchical} use CD score \citep{murdoch2018beyond} as a joining metric in the agglomerative clustering procedure;
\citet{chen2020generating} recursively divides large text spans into smaller ones by detecting feature interaction.
However, previous work requires only adjacent clusters to be grouped as a new cluster, which we denote as \textbf{the connecting rule}.
With the connecting rule, generated clusters will always be continuous text spans in the input text.
While consistent with human reading habits, we argue that the connecting rule as an additional prior may undermine the ability to faithfully reflect the model decision process.
The concerns are summarized as follows:

First, modern NLP models such as BERT \citep{devlin2018bert} and GPT \citep{radford2018improving, radford2019language} are
almost all transformer-based, using self-attention mechanisms \citep{vaswani2017attention} to build word relations. Since all word relations are calculated parallelly in the self-attention mechanism, connecting rule is inconsistent with the base working algorithms of these models.

Second, unlike the toy sample in Figure \ref{hierarchical attribution}, NLP tasks are becoming increasingly complex, often requiring the joint reasoning of different parts of the input text \citep{chowdhary2020natural}.
For example, Figure \ref{our method} shows an sample from natural language interface (NLI) task\footnote{NLI is a task requiring the model to predict whether the premise entails the hypothesis, contradicts it, or is neutral.}, in which \emph{`has a'} and \emph{`available'} are the key combinatorial semantics to make the prediction. 
However, hierarchical explanations with the connecting rule can not identify this compositional information but only can build relations  ntil the whole sentence is regarded as one cluster.


To this end, we propose to generate hierarchical explanations without the connecting rule and introduce a framework for generating hierarchical clusters, which produces hierarchical clusters by recursively detecting the strongest interactions among clusters and then merging small clusters into bigger ones.
Compared to previous methods with connecting rules, our method can provide compositional semantics information of long-distance spans.
We build systems based on two classic attribution methods: LOO \citep{lipton2018mythos} and LIME \citep{ribeiro2016should}. Experimental results and further analysis show that our method can capture higher quality features for reflecting model predicting than existing competitive methods.


\begin{figure}[htb]
	\centering
	\includegraphics[width=0.95\linewidth]{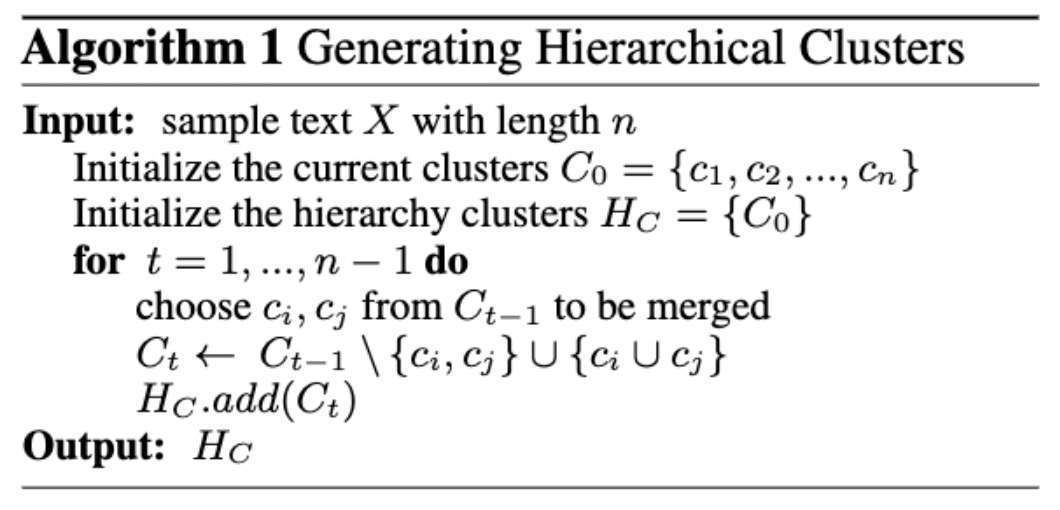}
	
\end{figure}

\begin{table*}[t!]
\centering
\begin{tabular}{l|cccccccccc}
\toprule[0.8pt]

\multirow{3}{*}{Method/Dataset} & \multicolumn{4}{c}{\underline{ \ \ \ \ \ \ \ \ \ \ \ \ \ \ \ \ \ SST-2 \ \ \ \ \ \ \ \ \ \ \ \ \ \ \ \ \ }} & \multicolumn{4}{c}{\underline{ \ \ \ \ \ \ \ \ \ \ \ \ \ \ \ \ \ MNLI \ \ \ \ \ \ \ \ \ \ \ \ \ \ \ \ \ }} & \multirow{3}{*}{\ \ avg} 
\\
&  \multicolumn{2}{c}{ \ \ \ \ \ AOPC$_{pad}$ \ \ \ \ \ } & \multicolumn{2}{c}{ \ \ \ \ \ AOPC$_{del}$ \ \ \ \ \  }  & \multicolumn{2}{c}{ \ \ \ \ \  AOPC$_{pad}$ \ \ \ \ \  } & \multicolumn{2}{c}{ \ \ \ \ \  AOPC$_{del}$ \ \ \ \ \  } 
\\
& $\emph{10\%}$ & $\emph{20\%}$  &    $\emph{10\%}$   & \multicolumn{1}{c}{$\emph{20\%}$} & $\emph{10\%}$ & $\emph{20\%}$  &    $\emph{10\%}$   & $\emph{20\%}$ \\
\midrule[0.4pt]

LOO  & 34.8 & 43.3 & 34.6 & 42.0  & 64.5 & 65.8 & 66.5 & 68.2 &   \ \   52.5 \\

 

L-Shapley  & 31.9 & 41.0 & 38.8 & 45.6   & 62.1 & 67.4 & 69.2 & 71.8 &   \ \   53.5 \\



LIME  & 32.8  & 46.8 & 36.6 & 49.7 & 67.3 & 71.1 & 70.1 & 75.3 &  \ \  56.2 \\
 
ACD$^\diamondsuit$   & 31.9 & 38.3 & 31.1 & 39.0   & 60.5  & 61.4 & 59.5 & 61.1  &  \ \   47.9   \\

HEDGE$^\diamondsuit$    & 34.3  & 46.7 & 34.0 & 44.1 & 68.2 & 70.9 & 68.3 & 70.9  &  \ \   54.7 \\

\midrule[0.4pt]

\rowcolor[gray]{0.9}
our HE$^\diamondsuit$  & 44.2 & 60.6 & 42.0 & 53.7  
 & 75.8 & 76.3 & 74.6 & 74.8 &   \ \ 62.8  \\
 


 

\bottomrule[0.8pt]
\end{tabular}
\caption{AOPC(10) and AOPC(20) scores of different explain methods in on the SST and MNLI datasets. $\diamondsuit$ refers to method with hierarchical structure. ${del}$ and ${pad}$ refer to different modification strategies in AOPC.}
\label{result1}
\end{table*}

\section{Method}

\subsection{Generating Hierarchical Clusters}

For a classification task, let let $ X = (x_{1},...,x_{n})$ denote a sample with $n$ words, and $c$ denotes a word cluster containing a set of words in $X$.
Algorithm \ref{Algorithm} describes the whole procedure of
hierarchical clusters.
With current clusters $C_{t}$ initialized with each $x$ as a cluster at step 0, the algorithm will choose two clusters from $C_{t}$ and merge them into one cluster in each iteration. After $n-1$ steps, all words in $X$ will be merged as one, and $C_{t}$ in each time step can constitute the final hierarchical clusters.

As shown in algorithm \ref{Algorithm}, to perform the merge procedure, we need to decide which cluster to be merged for the next step.
In this work, we choose clusters by finding the maximum interaction between clusters, 
 which can be formulated as the following optimization problem:
\begin{equation}
\begin{aligned}
\mathop{max}\limits_{c_{i},c_{j} \in C} \  \mathop{\phi} \ \mathop{(} \mathop{c_{i}}, \mathop{c_{j}} \mathop{|} \mathop{C},  \mathop{)}.
\label{1}
\end{aligned}
\end{equation}
$\phi(c_{i},c_{j}|C)$ defines the interaction score given current clusters $C$.

\subsection{Detecting Cluster Interaction}
Different previous work calculate interactions between two words/phrases according to model predictions, we calculate interactions between two clusters considering the influence of one cluster on the explanations of the other cluster.
Given an attribution algorithm $Algo$, 
quantified interaction score between $c_{i}$ and $c_{j}$ can be calculate as follows:
\begin{equation}
\begin{aligned}
\phi_{ij} &= abs(Algo(c_{i}) - Algo^{-c_{j}}(c_{i})) \\ & + abs(Algo(c_{j}) - Algo^{-c_{i}}(c_{j})) 
\label{2}
\end{aligned}
\end{equation}
where $Algo^{-c_{j}}(c_{i})$ denote the attribuition score of $c_{i}$ in the condition $c_{j}$ is marginalized.

Especially, previous work \citep{chen2020generating} generates hierarchical attributions by defining the following interaction score:
\begin{equation}
\begin{aligned}
\phi_{ij} &= E[f(x)|C \textbackslash (c_{i} \cup c_{j})]   - E[f(x)|C \textbackslash c_{i}] \\ &-  E[f(x)|C \textbackslash c_{j}]  + E[f(x)|C]
\label{3}
\end{aligned}
\end{equation}, 
where $f(x)|C$ denotes the mode prediction on $C$.
We demonstrate this definition is a special case of formula \eqref{2}, and the detailed proof are provided in the appendix.

\subsection{Attribution Algorithm.}
We use Leave-one-out (LOO) \citep{lipton2018mythos}  and LIME \citep{ribeiro2016should} as the basic attribution algorithms to build our systems, denoted as HE$_{loo}$ and HE$_{lime}$. 
LOO assigns attributions by the probability change on the predicted class after erasing a word.
Though the philosophy of LOO is simple, it is competitive, and only a forward pass is needed computationally.
LIME estimates word contribution to a label class by learning a linear approximation of the model's local behavior. LIME is a very powerful method to gain the attribution scores of the input text.  
Compared to LOO,  LIME can produce higher quality explanations but requires more computation time.



\section{Experiment}

\subsection{Dataset and Models.}
We adopt two text-classification datasets: binary version of Stanford Sentiment Treebank (SST-2) \citep{socher2013recursive}  and MNLI tasks of the GLUE benchmark \citep{wang2018GLUE}).
SST-2 is a Sentiment Classification task with binary sentimental labels;
MNLI is a Natural Language Inferencing (NLI) task that requires the model to predict whether the premise entails the hypothesis, contradicts it, or is neutral.
We build the target model with BERT$_{base}$ \citep{devlin2018bert} as encoder, achieving 91.7\% and 83.9\% accuracy on SST-2 and MNLI.

\subsection{Evaluation Metrics. \ }
Since the reasoning process of deep models
is inaccessible, researchers design various evalu-
ation methods to quantitative evaluate the faithfulness of explanation of deep models, among which the most widly used evaluation metric is the area over the perturbation curve (AOPC).\footnote{Since different modification strategies might lead to different evaluation results on AOPC \citep{ju-etal-2022-logic},  we evaluate with both modification strategies $del$ and $pad$. $del$ modify the words by deleting them from the origianl text directly while $pad$ modify the words by replacing them with [`PAD'] tokens. Moreover, since we didn't introduce new strategy to get attribution scores, it avoid the risk of unfair comparisons due to customized modification strategies.}
By  modifying the top k\% words, AOPC calculates the average change in the prediction probability on the predicted class over all test data as
follows,
\[
AOPC(K) = \frac{1}{N}\sum_{i=1}^{N}\left \{p(\hat{y}|x_{i}) - p(\hat{y}|\tilde{x}_{i}^{(k)}) \right \}
\]
where $\hat{y}$ is the predicted label, $N$ is the number of examples, $p(\hat{y}|·)$ is the probability on the predicted class, and $\tilde{x}_{i}^{(k)}$ is modified sample. Higher AOPCs is better, which means that the features chosen by attribution scores are more important.
For a cluster containing multiple words, we distribute attribution score of this cluster to each word equally.
If a cluster is chosen for calculating the AOPC score, other clusters containing this cluster will not be chosen.
We strictly guarantee that the number of words chosen for each evaluated method is the same.
If the chosen cluster contains more words than needed, we pick words individually by their attribution scores.

\subsection{Result Compared to Other Methods}

We compare our HE ith several competitive baselines, among which LOO, Shapley \citep{chen2018shapley}, and LIME are non-hierarchical, and ACD \citep{singh2018hierarchical} and HEDGE \citep{chen2020generating} are hierarchical methods.
As shown in Table \ref{result1}, except for LIME, other baselines (hierarchical or not) have no obvious improvement compared to LOO.
In contrast, our LOO-based hierarchical explanation outperforms LOO by an average of over 7\%. Moreover, our LIME-based hierarchical explanation outperforms LIME and achieves the best performance.
Results in Table \ref{result1} demonstrate the effectiveness of our approach in generating hierarchical explanations based on existing attribution algorithms.

\subsection{Result of Ablation Experiment}

Though HE$_{loo}$ and HE$_{lime}$ outperform their basic attribution algorithms, they provide attribution scores of more word clusters than the non-hierarchical method.
To further demonstrate the effectiveness of our approach and the negative influence of the connecting rule, we conduct an ablation experiment with two special baselines: HE-random and HE-connecting.
HE-random and HE-connecting are modified from our proposed method with the following modifications:
HE-random merges clusters randomly in each iteration and
HE-connecting merges connecting clusters with the largest non-additive score.





\begin{figure}[t!]
	\centering
	\includegraphics[width=0.85\linewidth]{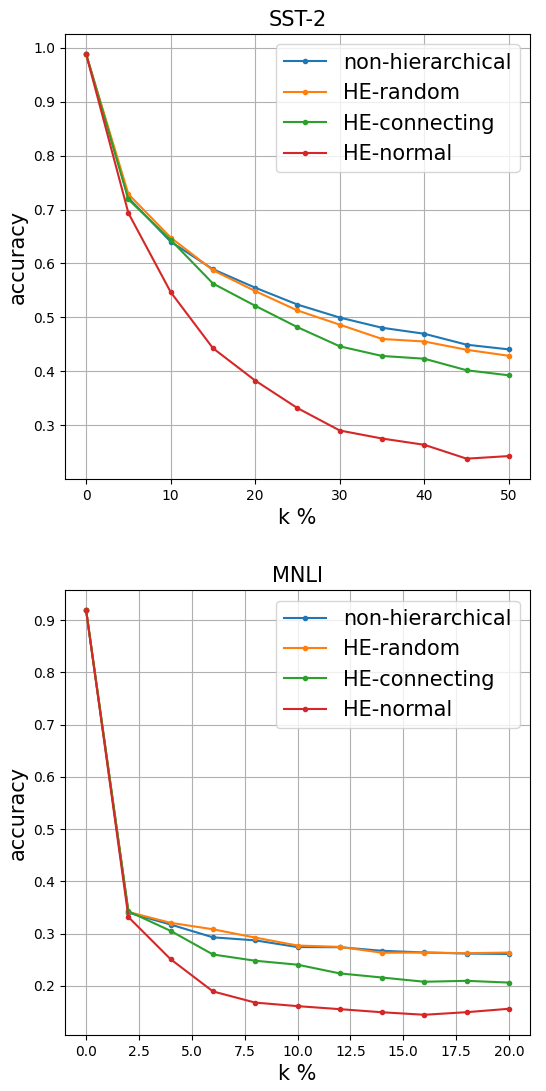}
	\caption{Model accuracy after padding the top k\% words according to the explanation.}
	\label{result2}	
\end{figure}

Figure \ref{result2} shows the rate of decline in model accuracy after padding the top $k\%$ words according to the explanation. The faster the accuracy declines, the better the explanation, demonstrating that the selected words are important.
\footnote{Essentially same as AOPC$_{pad}$, equals to original accuracy minus  AOPC$_{pad}$ score.}
As shown in Figure \ref{result2}, 
HE-random on SST-2 slightly outperforms non-hierarchical explanations on some $k$ but has no improvement on MNLI. We assume that this is because the sentences of SST samples are relatively short, randomly combined clusters have greater chance of containing meaningful combinations.\footnote{The number of all possible combinations in each iteration is approximately equal to the square of the sentence length.}
The HE-connecting and HE-normal have significant improvements over non-hierarchical method and HE-random, demonstrating the effectiveness of our approach in building hierarchical explanations. Moreover, the HE-normal outperform He-connecting consistently in all dataset and all $k$, demonstrating the argumentation that the connecting rule limits the hierarchical explanation to faithfully reflect model predicting.

\section{Conclusion}
In this paper, we propose to generate hierarchical explanations without the connecting rule and introduce a framework for generating hierarchical clusters. 
 We build systems with LOO and LIME attribution algorigm on two text classification benchmarks. Experimental result compaerd with several competitive baseline
methods and the result of ablation Experiment demonstate the superiority of our method and the argumet of  the connecting rule organize and explain better response model predictions..

\section*{Limitation}
In this work, we propose quantifying interaction between word sets by the influence of one set on the explanations of the other set (Formula \ref{2}).
In theory, this method can be applied with any attribution algorithm $Alog$.
However, since our target is building hierarchical attributions without the connecting rule, we need to calculate $n^{2}$ times attribution score for each sample (seen Section \ref{Complexity}). Thus, we use a simple $Alog$ in our experiment, making our experiment more computational economical than explanation methods that require sampling, such 
as LIME, Sample-Shapley and HEDGE.
However, the computational cost of Formula \ref{2} will increase linearly with more complex $Alog$, which limits the use of some computationally $Alog$s as our basic attribution methods.
For example, if we use LIME or Sample-Shapley as the baic $Alog$s to building systems, we need do sampling for calculating each interaction score, which is very computation costing.   



\bibliography{anthology,custom}
\appendix

\begin{figure*}[htb]
	\centering
	\includegraphics[width=0.99\linewidth]{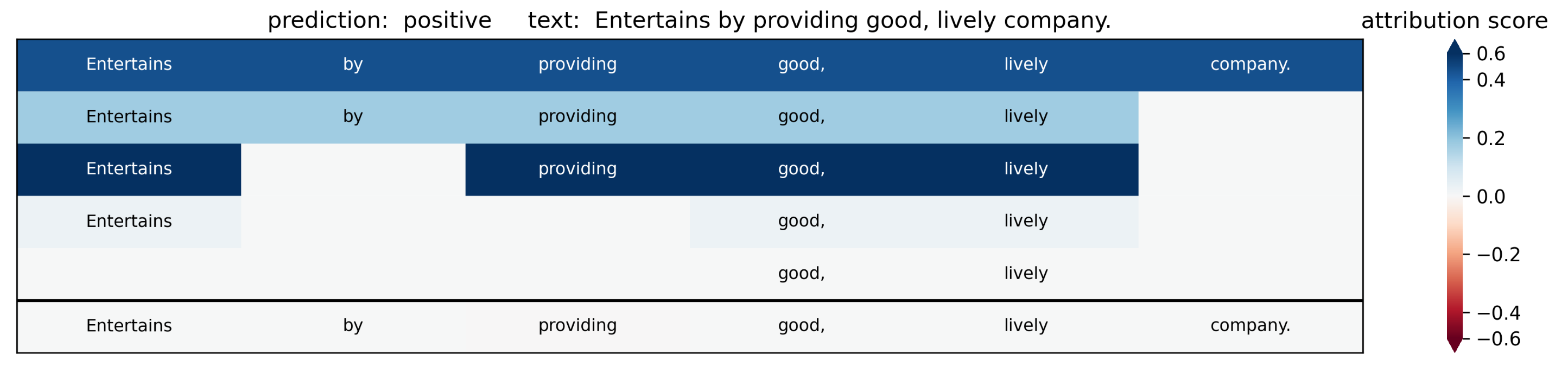}
	\caption{An example of visualized hierarchical
attributions without the connecting rule. Positive attribution score indicates supporting the model prediction while negative attribution score indicates opposing  model prediction. }
	\label{exa}	
\end{figure*}

\section*{Appendix}

\section{Experiment details}
For the SST dataset, we test all samples from dev set. For the MNLI dataset, we tested on a subset with 1000 samples (the first 500 samples from dev-matched and the first 500 samples from dev-mismatched.) due to computation costs.
Our implementations are based on the Huggingface’s transformer model hub (\url{https://github.com/huggingface/transformers}), and we use its default model architectures for corresponding tasks. \label{web}

\section{Visualization of Hierarchical
Attributions}
Since the generated set of words is discontinuous without connecting rule, we only represent the newly generated set and the corresponding attributions at each layer. For example, Figure \ref{exa} shows an example of the visualization of hierarchical
attributions. The bottom row shows the attribution score with each word as a set (non-hierarchy); The second-to-last row indicates the \{\emph{good}\} and \{\emph{lively}\} are merged togather and form a new set: \{\emph{good}, \emph{lively}\}; Similarly, the third-to-last line indicates the \{\emph{Entertains}\} and \{\emph{good}, \emph{lively}\} are merged togather.

We provide visualizations of all hierarchical attributions used for evaluation in the supplementary material (2000 samples). 
Moreover, for the convenience of reading, we also select some short-length examples and put them in this section. The visualization of hierarchical
attributions show that the HE can not only get significant improvement on quantitative evaluation but also are easy to understanding for humans.

\section{Formula Proof}
This section gives the proof that Formula \ref{3} is special case of our proposed method (Formula \ref{2}).
Following the theory in LOO \citep{lipton2018mythos}, the attribution of a subset in the input text can be calculate by the probability change on the predicted class after erasing this set. Thus, the attribution of $c_{i}$ can be calculate by $E[f(x)|C] - E[f(x)|C \textbackslash c_{j}]$, denoted as: 
\begin{equation}
\begin{aligned}
Algo(c_{i}) = E[f(x)|C] - E[f(x)|C \textbackslash c_{j}].
\nonumber
\end{aligned}
\end{equation}
Moreover, erasing can been seen as a specific method for marginalizing a set in Formula \ref{2}. Thus,
\begin{equation}
\begin{aligned}
Algo^{-c_{j}}(c_{i})) = (E[f(x)|C] - E[f(x)|C \textbackslash c_{i}])^{-c_{j}} \\
 =  E[f(x)|C \textbackslash c_{j}] - E[f(x)|C \textbackslash (c_{i} \cup c_{j})].
\nonumber
\end{aligned}
\end{equation}
Thus, it can be deduced that under such $Algo$ and marginalization method,
\begin{equation}
\begin{aligned}
(Algo(c_{i}) - Algo^{-c_{j}}(c_{i})) &= 
(Algo(c_{j})  \\ &- Algo^{-c_{i}}(c_{j})),  
\nonumber
\end{aligned}
\end{equation}
which both equal to Formula \ref{3}:
\begin{equation}
\begin{aligned}
\phi_{ij} &= E[f(x)|C \textbackslash (c_{i} \cup c_{j})]   - E[f(x)|C \textbackslash c_{i}] \\ &-  E[f(x)|C \textbackslash c_{j}]  + E[f(x)|C]
\nonumber
\end{aligned}
\end{equation}.
Note that since Formula \ref{3} does not take the absolute value of the interaction score.
This strategy will leads to the interactions of reducing each other's attributions are ignored, which does not consistent with the definition of interaction score.




\begin{figure*}[htb]
	\centering
	\includegraphics[width=0.99\linewidth]{ 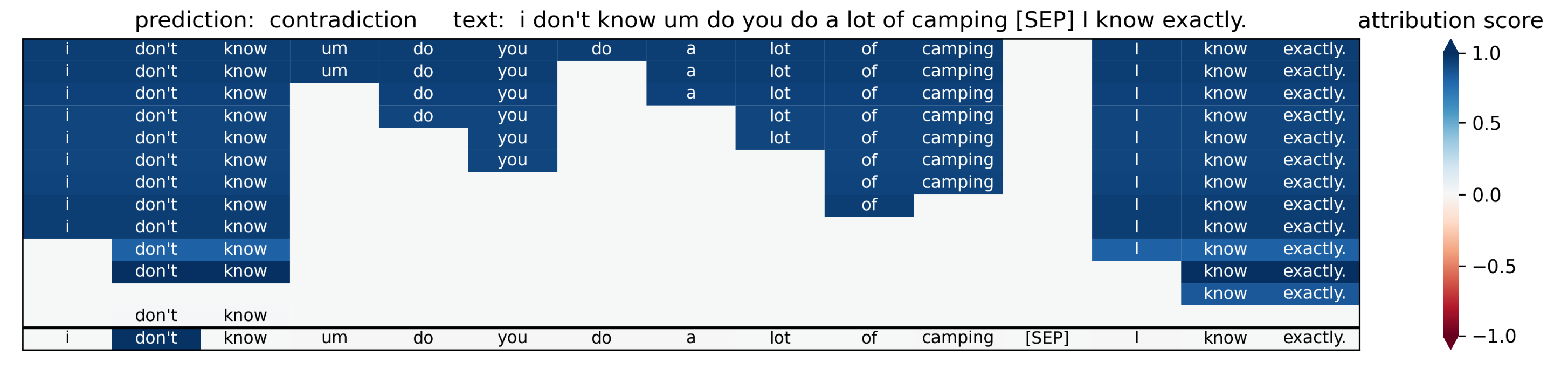}
	\caption{HE for 4th sample in MNLI dev set.  \emph{don't know} and \emph{know exactly} are recognized at first and then the combination of them are recognized as `contradiction'.}
	\label{e3}	
\end{figure*}

\begin{figure*}[t!]
	\centering
	\includegraphics[width=0.99\linewidth]{ 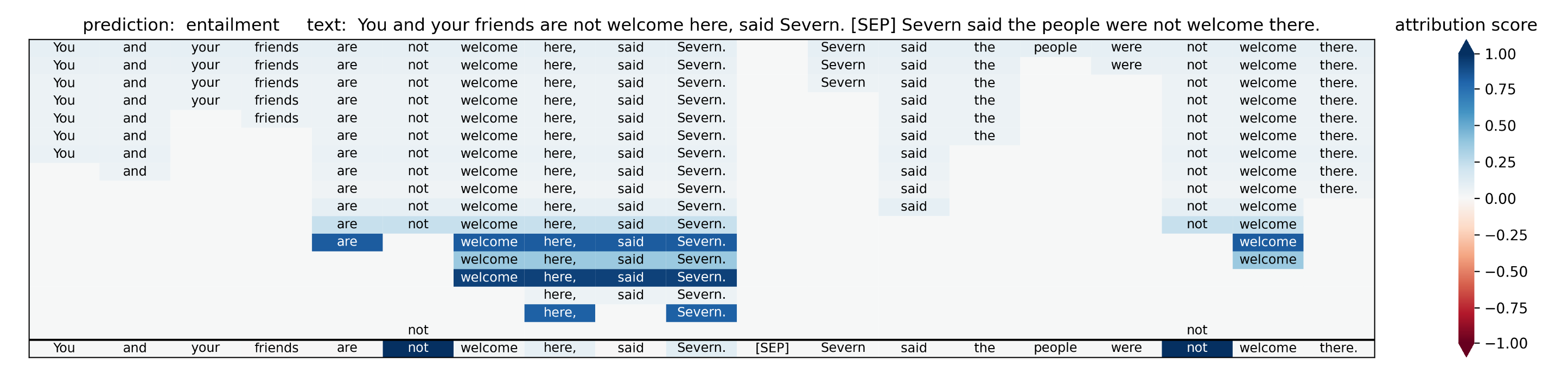}
	\caption{HE for 13th sample in MNLI dev set.  \emph{not} as a transition are recognized as both sentences first. Then the HE capture the \emph{welcome here} and \emph{welcome}, which have the same meaning. Thus the model make the prediction: `entailment'. }
	\label{e4}	
\end{figure*}

\begin{figure*}[t!]
	\centering
	\includegraphics[width=0.99\linewidth]{ 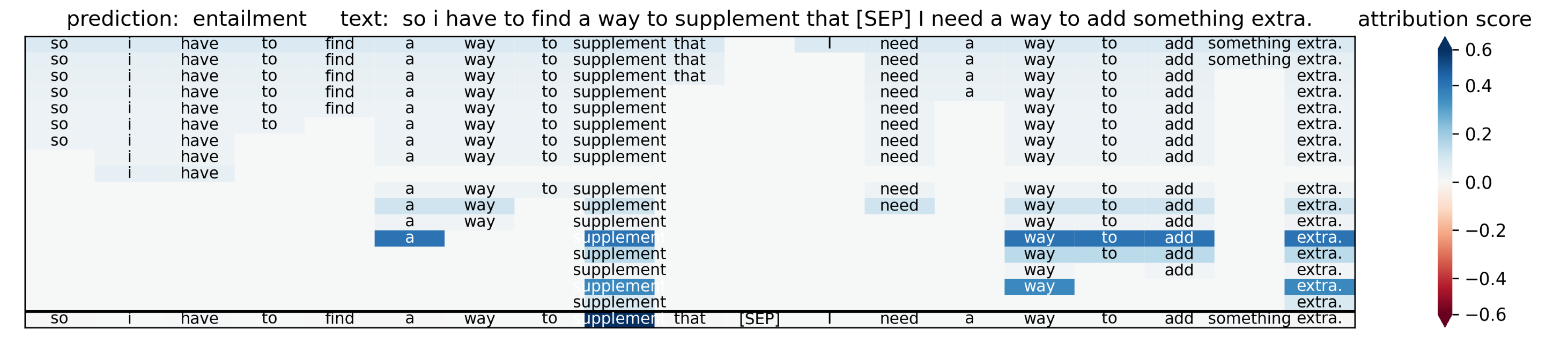}
	\caption{HE for 16th sample in MNLI dev set. 
     \emph{a supplement}, \emph{way to add} and \emph{extra} are recognized at the key feature to make the prediction: `entailment'.}
	\label{e5}	
\end{figure*}

\begin{figure*}[t!]
	\centering
	\includegraphics[width=0.99\linewidth]{ 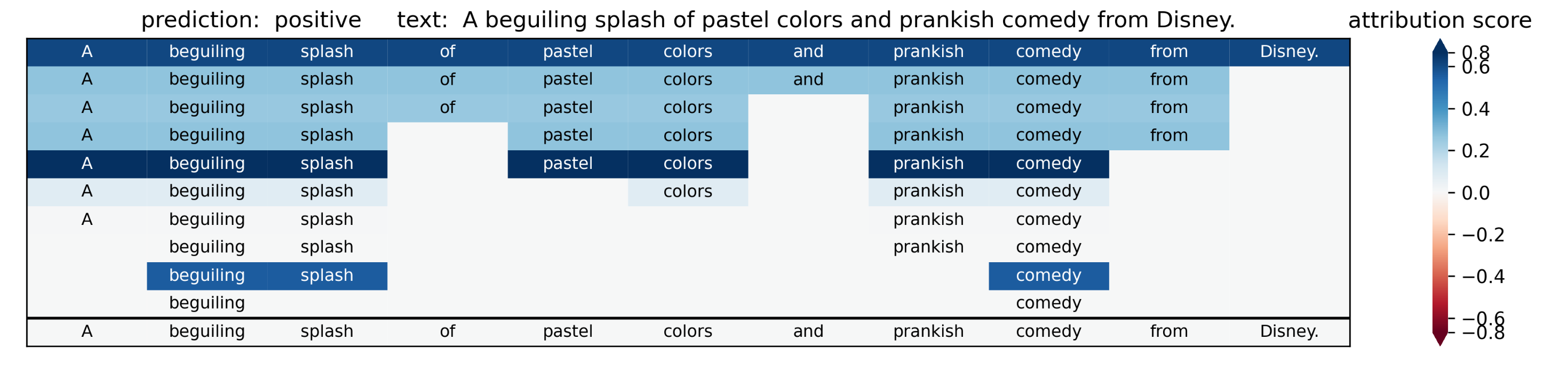}
	\caption{HE for 12th sample in SST dev set. HE shows \emph{beguiling splash} and \emph{comedy} are merged together and support the prediction: `positive', and the support improved when the \emph{pastel colors} are merged in too.}
	\label{e1}	
\end{figure*}

\begin{figure*}[t!]
	\centering
	\includegraphics[width=0.99\linewidth]{ 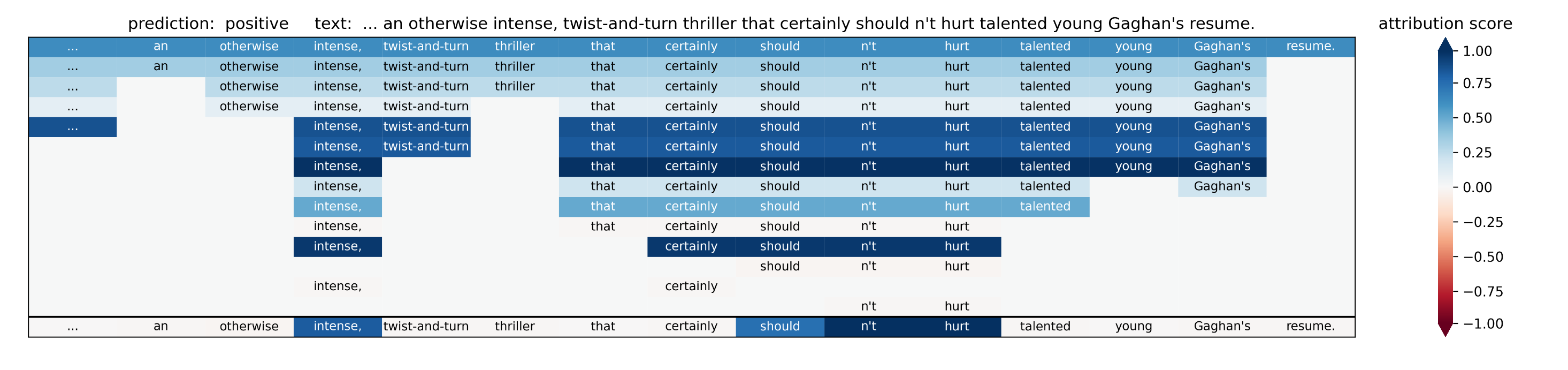}
	\caption{HE for 17th sample in SST dev set. \emph{intense} and \emph{certainly} are merged together first as a emphasis. Then the \emph{should n't hurt} are merged together. Then the emphasis and \emph{should n't hurt} are merged together.}
	\label{e2}	
\end{figure*}

\section{Experimental Computation Complexity}
\label{Complexity}
Since we need to choose the maximum set interaction in each iteration. For the step 1, we need to calculate the interaction score between between each two sets. In other step, we need to calculate the interaction scores between the new generated set and other sets. In total, we need to calculate $C_{n}^{2} + (n-2) + ,..., 1 \approx n^{2}$ times interaction score. where $n$ refers to the sequence length of the input text. In our experiment, calculate an interaction score is comparable to three forward pass through the network. Thus, the total computation complexity of building our hierarchical attributions for a sample is $3n^{2}$ forward pass through the network.
Note that through record the model prediction during every iteration, the computational complexity can be reduced by about half.

Compared to explanation methods that do not require sampling, such as CD  \citep{murdoch2018beyond} and LOO \citep{lipton2018mythos}, our experiment requires more computation. Compared to explanation methods that require sampling, such 
as LIME \citep{ribeiro2016should}, Sample-Shapley \citep{strumbelj2010efficient} and HEDGE \citep{chen2020generating}, which has a full sample space of $2^{n}$ and often need to sample thousands of samples, our experiment is more computational economical.
Note that we use a simple algorithm to calculate attribution score in our experiment. The computational cost will increase linearly with more complex attribution algorithm. (seen section Limitation)


\end{document}